# Broiler-Net: A Deep Convolutional Framework for Broiler Behavior Analysis in Poultry Houses

Tahereh Zarrat Ehsan [1], Seyed Mehdi Mohtavipour [2]

[1] zarrat.ehsan@gmail.com
[2] mehdi_mohtavipour@elec.iust.ac.ir
[1] School of Electrical Engineering, University of Guilan, Rasht, Iran
[2] School of Electrical Engineering, Iran University of Science and Technology, Tehran, Iran

**Abstract**. Detecting anomalies in poultry houses is crucial for maintaining optimal chicken health conditions, minimizing economic losses and bolstering profitability. This paper presents a novel real-time framework for analyzing chicken behavior in cage-free poultry houses to detect abnormal behaviors. Specifically, two significant abnormalities, namely inactive broiler and huddling behavior, are investigated in this study. The proposed framework comprises three key steps: (1) chicken detection utilizing a state-of-the-art deep learning model, (2) tracking individual chickens across consecutive frames with a fast tracker module, and (3) detecting abnormal behaviors within the video stream. Experimental studies are conducted to evaluate the efficacy of the proposed algorithm in accurately assessing chicken behavior. The results illustrate that our framework provides a precise and efficient solution for real-time anomaly detection, facilitating timely interventions to maintain chicken health and enhance overall productivity on poultry farms.
Github: https://github.com/TaherehZarratEhsan/Chicken-Behavior-Analysis

## 1. Introduction

Chicken meat is a significant source of protein for people, and its consumption has increased considerably in the last 30 years [1]. According to recent research, the global poultry market is expected to grow from $352.02 billion in 2022 to $487.39 billion in 2027, with a compound annual growth rate of 7.6%. This growth places formidable challenges on farmers to meet the escalating demand. Traditionally, farmers have relied on manual observation for monitoring chicken conditions, a time-consuming and error-prone task due to the sheer number of chickens. However, the decreasing cost of technological devices [2] and introduction of powerful Deep Neural Network (DNN) [3] have facilitated the widespread adoption of monitoring systems in commercial farms. Leveraging artificial intelligence (AI), these systems autonomously monitor chickens, offering farmers an efficient means to manage poultry houses and enhance profitability.

The rapid growth of chicken production also increases the risk of spreading diseases, leading to economic losses and posing threats to human health. Therefore, designing an automatic monitoring tools to improve farm welfare is an important topic in the computer vision community. In [4], bird postures are analyzed using skeleton shapes to detect sickness, employing an ellipse segmentation algorithm and handcrafted features for classification. Support vector machines are then trained to classify broilers as healthy or sick. Similarly, in [5], contours are extracted, and the distance between the highest point of the chicken body and the camera sensor is computed to classify birds as standing or lying. Another method proposed in [6] to detect sick chickens in caged farms by segmenting chicken body parts using active contour techniques to obtain the heads. The time period of eating and drinking is calculated, and chickens with slow behavior are selected as sick. In [7], chickens are equipped with wearable IoT sensing devices to obtain their behavior pattern over time. Generative Adversarial Network (GAN) is utilized to produce synthetic data and increase the size of the dataset. Several machine learning models are trained on a combination of real and synthetic data to classify samples as sick or healthy. A monitoring system based on image processing technique is proposed in [8] to predict the weight of the chicken. This system utilized a watershed segmentation algorithm to segment the images and handcrafted descriptors to capture weight information. Bayesian Neural Network (BNN) is trained on these features to predict the weight. In [9], a model was developed to analyze the drinking and feeding behavior of broilers using machine learning models to estimate the total number of birds at drinkers and feeders. Additionally, [10] proposed a neural network-based method to detect the stunned state in chickens using a Convolutional Neural Network (CNN) model. Another CNN model is designed in [11] to classify chicken behaviors to six classes of standing, walking, running, eating, resting and preening. A model is designed to estimate chicken pose and a naïve Bayes classifier is trained on the chicken pose to categorize behaviors. Similarly, another behavior classification method is presented in [12] which classify behaviors to three classes of eating, sleeping and waling. Chicken trajectories are obtained and handcrafted features are extracted from the trajectories. Several machine learning models including logistic regression and naïve base classifier are used for classification.

A tracking method is proposed in [13] to monitor chickens in consecutive frames of the video. A regression neural network is developed to find the location of the chicken in the next frame based on the chicken bounding box from the previous frame. Another tracking tool is developed in [14] to detect and track chickens across the video. You Only Look Once (YOLO) model is utilized in this work to detect chicken in each frame and kalman filter is used for tracking. Similarly, a tracking method is proposed in [15] to obtain chicken trajectory over time. In [16], a combination of YOLO object detector and deep sort tracking algorithm is utilized to obtain a mobility assessment framework. Authors in [17] proposed a segmentation network to separate chicken from the background in the poultry house. A multi-scale encoder decoder network with attention module is designed to focus on important features for segmentation. Another YOLO model is presented in [18] to detect cage-free chicken on the litter floor. Similarly, a method based on YOLO is presented in [19] to detect chicken face from the image. Generative Adversarial Networks (GAN) are employed for data augmentation, enhancing the dataset's diversity. The YOLO (You Only Look Once) architecture is refined to achieve improved accuracy in detecting small-size targets. In [20], another YOLO model is designed for laying and bath-dusting behavior classification. Finally, in [21], different light colors and temperature environments are created and chickens behaviors in these environments are analyzed with YOLO model for more than 648 hours to assess their environment preference. Table 1 provides details of artificial intelligence-based methods designed to aid farmers in managing poultry houses. As can be seen, all the previous works are evaluated using private datasets. To the best of our knowledge, there is no public dataset for analyzing the abnormality in chickens.

This paper introduces an innovative framework designed to identify abnormal behaviors in chickens, with a specific emphasis on inactivity and huddling. Engineered for real-time operation, the framework is well-suited for on-the-edge devices. Inactivity detection holds particular significance as it can serve as an early indicator of chicken sickness. Sick chickens often display sedentary behavior and limited movement within the poultry house. Furthermore, huddling behaviors are also detected, where chickens gather closely together. In severe instances, chickens may pile on top of each other, potentially leading to illness and fatalities. To mitigate these issues, an artificial intelligence-based model is developed to promptly identify and alert farmers to these abnormal behaviors. The ultimate objective of this model is to substantially enhance the health and welfare of chickens, thereby elevating overall farm productivity.

Table 1. Related works based on computer vison and machine learning techniques

| Reference | Application | Dataset availability |
| --- | --- | --- |
| [4-7] | Sick chicken detection | Private |
| [8] | Chicken body weight estimation | Private |
| [9-12], [20, 21] | Chicken behavior classification | Private |
| [18] | Chicken detection | Private |
| [19] | Chicken face detection | Public |
| [13], [15] | Chicken tracking | Private |
| [14], [16] | Chicken detection & tracking | Private |
| [17] | Chicken segmentation | Private |

Broiler-Net: A Deep Convolutional Framework for Broiler Behavior Analysis in Poultry HousesThe remainder of this paper is as follows, section 2 provides a detailed discussion on the dataset and the proposed framework. Section 3 describes the results obtained, offering insights into the outcomes of the study. Finally, the concluding remarks are presented in the last section of this paper

## 2. Material and methods
### 2.1 Dataset

According to our survey, the lack of a public dataset for chicken welfare analysis in poultry houses presents a significant challenge in this field, limiting researchers' ability to work on this topic. Additionally, the number of research papers on chicken behavior analysis lags considerably behind other computer vision domain such as human behavior analysis [22-24] and human abnormal behavior detection [25-27].

In this paper, we address three distinct problems: chicken detection, inactivity detection, and huddling detection. To collect the dataset, we gathered videos of chickens in cage-free poultry farms from the Google search engine. For chicken detection, each video was divided into frames and frames labeled using the LabelImg software as it is shown in figure 1. A bounding box was manually drawn for each chicken, and XML files were produced which contains the bounding box coordinates and corresponding labels. These XML files and frames were then used to train the chicken detection module. Since the training dataset was not large-scale, we utilized data augmentation techniques such as vertical and horizontal shift, rotation, and brightness changes to increase the size of the training dataset. This allowed the model to learn to recognize different variations of the samples and improve training. The details of the dataset can be found in Table 1, which includes both the original and augmented datasets consisting of 2080 and 10400 images, respectively. The dataset was split into 90% for training, 10% for validation, and 10% for testing.

For abnormality detection, frames were classified as depicting huddling when more than 10 chickens were observed closely congregating within a 100-pixel radius. Careful examination of the dataset allowed us to annotate 123 frames exhibiting the huddling condition. In the detection of inactive broilers, chickens were labeled as inactive if their movements were less than 20 pixels in 50 consecutive frames, resulting in the observation and manual annotation of a total of 71 inactive chickens within the dataset. Since huddling and inactivity detection do not need training in our framework, there was no need for data augmentation or the establishment of train, validation, and test splits. The dataset statistics are succinctly summarized in Table 2.

Table 2. Statistics of dataset

|  | Original dataset | Augmented dataset | train | validation | test |
| --- | --- | --- | --- | --- | --- |
| Chicken detection | 2080 | 10400 | 11232 | 1248 | 1248 |
| Huddling detection | 123 | - | - | - | - |
| Inactivity detection | 71 | - | - | - | - |

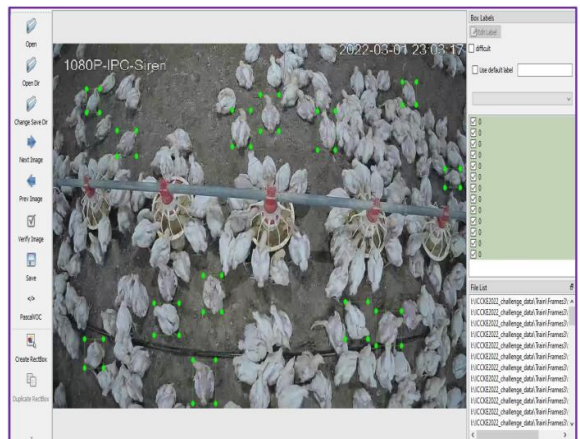

Figure 1. LabelImg environment for data annotation

## 2.2 Flowchart of the proposed framework

Figure 2 illustrates the flowchart of our proposed framework for detecting abnormal behavior in chickens. The framework comprises three key components: a chicken detection module, a chicken tracking module, and two abnormality detection modules. Firstly, the input video is divided into frames, and in each frame, chickens are extracted to identify huddling behavior. Subsequently, the movement of the chickens is tracked across consecutive frames to analyze their activity levels and detect inactivity abnormality. In the subsequent sections, we will provide a comprehensive discussion of each component.

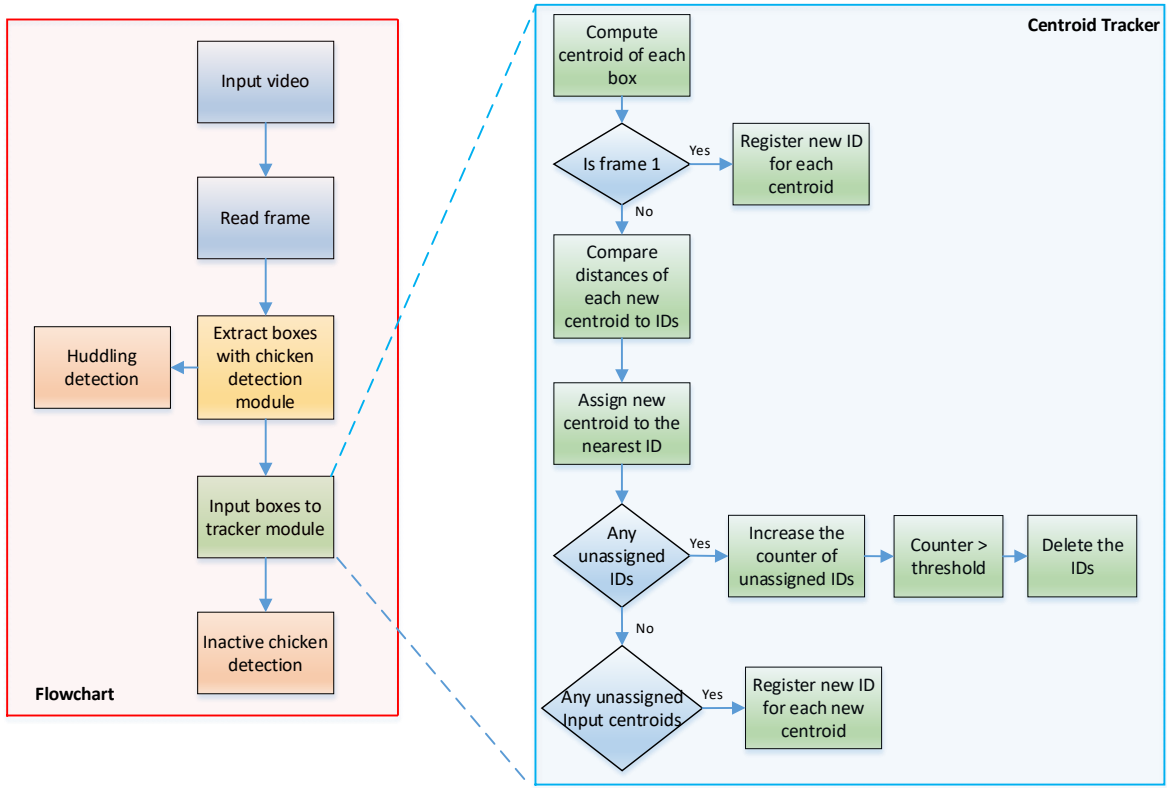

Figure 2. Flowchart of the proposed framework for abnormal chicken detection

## 2.3 Chicken detection module

Object detectors are composed of three main parts: the backbone, neck, and head. The backbone extracts features from the input image using convolutional layers. These layers extract features at different scales, ranging from low-level features like edges and contours to high-level features like shapes and object parts. Popular networks used for the backbone include DarkNet [28], EfficientNet [29] and ResNet [30]. The extracted features are then combined in the head part to obtain more informative features. Feature Pyramid Network (FPN) [31] and Path Aggregation Network (PANet) [32] are two well-known head methods is responsible for detecting objects based on the extracted features. Finally, non-maximum suppression (NMS) is used to remove duplicate objects.

In this work, as it is shown in figure 3, YOLO v4 network [33] is specialized for broiler detection. Since there is no broiler class in the COCO dataset on which YOLO v4 is trained, we use our own dataset for training. In order to specifically detect the broiler class, we have made modifications to the final layers of YOLO v4. This adaptation is crucial as YOLO v4 is originally designed to detect 80 different classes, whereas our research objective focuses solely on identifying broilers. By modifying the final layers, we ensure that our model is trained to accurately detect and classify broilers, while disregarding other classes. As our dataset is not large-scale, training YOLO from scratch is not feasible. Instead, we start with a pre-trained YOLO model and fine-tuned it on the broiler images. This approach allows the model to utilize previously learned features and incorporate new information without the need to train from scratch on a large-scale dataset.

CSPDarknet-53 which consists of 53 convolutional layers with Cross Stage Partial (CSP) blocks is utilized as the backbone. CSP divides the feature maps into two parts and computes the residual connection only in one part, which reduces computational complexity. It also improves gradient flow in the residual connections and enhances model convergence. Therefore, CSPDarknet-53 can extract valuable information with low computational costs. The extracted information is then fed to the Spatial Pyramid Pooling (SPP) layer, where three maxpooling layers with sliding window sizes of 5, 9, and 13 are applied to extract features at different scales. For example, a window size of 5 focuses on smaller objects while a window size of 13 pays more attention to larger objects. The outputs of the maxpooling layers are concatenated and passed to the neck part. In the neck part, PANet is used to enhance information flow through the pipeline. PANet employs a top-down and bottom-up path to propagate information in the model. As features pass through layers, the image resolution decreases and the network extracts more semantically complex features. PANet combines earlier and deeper layers to leverage both low-level and semantically rich features. Features at different scales are then fed to the final head part for detection, with each head focused on detecting objects of a specific scale.

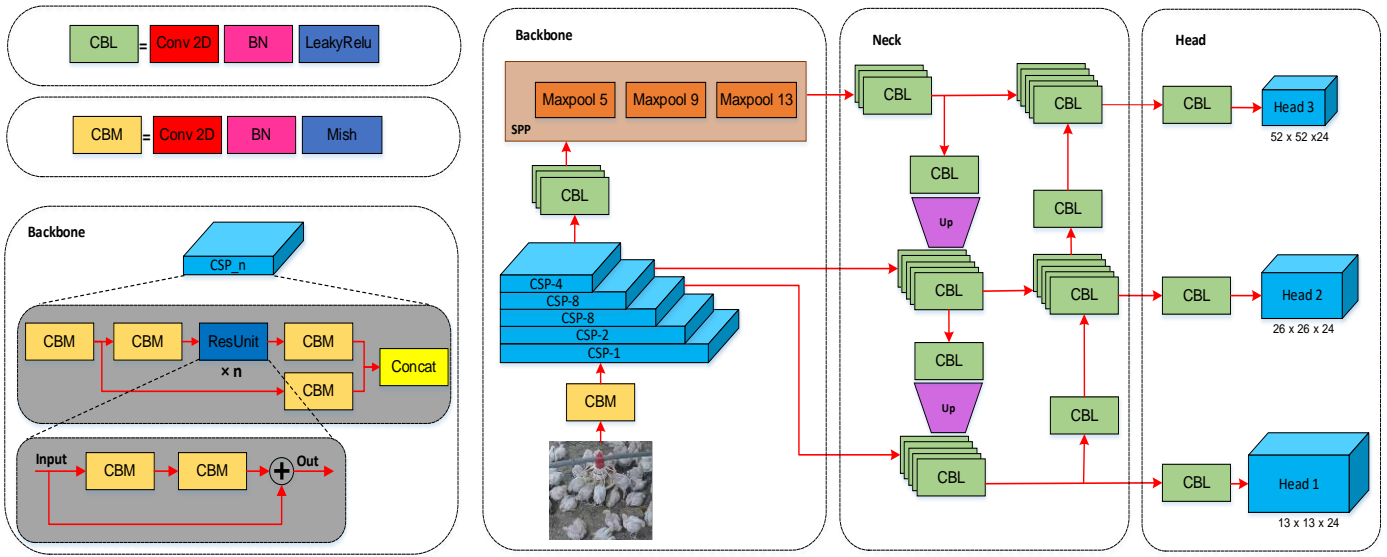

Figure 3. YOLO v4 architecture for chicken detection

## 2.4 Tracker module

To track broilers across video frames, the centroid tracker [34] is employed due to its speed and low computational resource requirement. The Centroid tracker is particularly suitable for running on edge devices with limited computational capabilities as it solely relies on Euclidean distance computation. As it is shown in figure 2, centroid tracker comprises three main components. Firstly, broilers are detected at each frame using the YOLO detector. This step allows us to identify the presence of broilers in every frame of the video. Next, the Euclidean distance between the newly detected broilers and the broilers present in the previous frames is computed. Finally, we assign the new broilers to the broilers with the lowest distance. By assigning each new broiler to its nearest counterpart from the previous frames, a consistent tracking system for individual broilers across the video frames is established. An illustrative example of the centroid tracker can be seen in the figure 4. In this example, ID 1, ID 2, and ID 3 represent the broilers detected at frame t. YOLO is then applied to frame t+1 to detect new broilers denoted by Unknown 1, Unknown 2, and Unknown 3. The Euclidean distance between each Unknown broiler and the IDs is computed, and the Unknown broilers are assigned to the nearest ID based on the distance calculation. This integration of YOLO v4 and the Centroid tracker allows to accurately detect and track broilers in video frames, providing valuable insights for broiler monitoring.

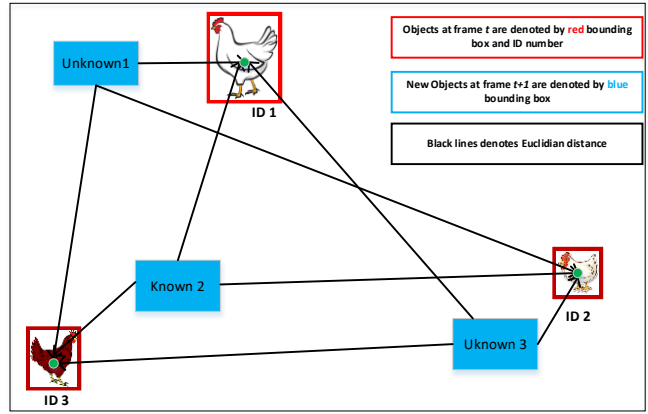

Figure 4. Broiler tracking idea

## 2.5 Huddling behavior module

In order to address the issue of huddling behavior in chickens, which can occur due to various factors such as cold environments, limited coop space, and potential threats, we have developed a model specifically designed to detect huddling behavior in cage-free houses. To achieve this, the Euclidean nearest neighbor search technique [35] is utilized, which allows to determine the total number of chickens within a fixed radius surrounding each detected object. A radius of 100 pixels around each object is chosen and the number of chickens present within this radius is calculated. If the total number of chickens exceeds a certain threshold, the frame is classified as huddling behavior. Our findings indicate that huddling behavior is predominantly observed when more than ten chickens are located within a radius of 100 pixels. Therefore, the threshold is set to 10 for the final evaluation of the model. By implementing this method, farmers are provided with a tool to identify huddling behavior in its early stages and take preventive measures to avoid potential losses. This approach enables proactive management of chicken welfare and aids in maintaining optimal living conditions for the flock.

## 2.6 Inactive broiler detection

To address the issue of identifying abnormal broilers in poultry houses, we propose a method based on measuring the activity level of chickens. Inactive chickens are often indicative of sickness, as healthier chickens tend to move more within the coop. To measure the activity level, the total displacement of chickens in consecutive frames is computed. In each frame, the chicken is detected and represented with a rectangular bounding box with four values: $x_{min}$, $y_{min}$, $width$ and $height$ which $x_{min}$ and $y_{min}$ are the x and y coordinate of the lower left corner of the rectangle. The displacement between two frames is then calculated using the following formula:

$$Displacement(i, i-1) = \sqrt{(Centroid_x(i) - Centroid_x(i-1))^2 + (Centroid_y(i) - Centroid_y(i-1))^2} \quad (1)$$

Where $Centroid_x(i)$ and $Centroid_y(i)$ are the x and y coordinate of the centroid of the bounding box in frame $i$ and computed as follows:

$$Centroid_x, Centroid_y = x_{min} + 0.5 width, y_{min} + 0.5 Height \quad (2)$$

The activity level is obtained by summing up the displacements over a specified number of consecutive frames (T):

$$Activity_{level} = \sum_{i=t}^{t+T} Displacement(i, i-1) \quad (3)$$

To identify inactive chickens, a threshold is applied to the activity level. Chickens with activity levels below this threshold are considered candidates for sickness and can be further analyzed by farmers to assess their health conditions. This method saves farmers significant time and effort compared to individually analyzing each chicken's health condition, thereby increasing productivity. In conclusion, the proposed method offers a practical and efficient approach to detect abnormal broilers based on their activity levels. By quickly identifying potentially sick chickens, farmers can take timely action to prevent further spread of disease and ensure the overall well-being of their flock

## 3 Results

In this section, the results of our proposed framework is presented. Recall, Precision, $F_1$ score and mean Average Precision (mAP) is utilized for evaluating the proposed framework. Recall is calculated as the number of correctly detected samples divided by the total number of samples:

$$Recall = \frac{TP}{TP + FN} \quad (4)$$

Precision measures the percentage of correct predictions out of the total number of detected samples:

$$Precision = \frac{TP}{TP + FP} \quad (5)$$

$F_1$ score is a measure of overall model performance:

$$F_1 score = \frac{2 \times Recall \times Precision}{Recall + Precision} \quad (6)$$

mAP is a metric for measuring the accuracy of the object detector. It is computed using the following equation:

$$mAP = \frac{\sum_{i=1}^{num\_classes} AP_i}{num\_classes} \quad (7)$$

Which $AP_i$ is the average precision for class $i$. Since the chicken detection problem only involves one class of chicken, mAP is equivalent to AP. mAP is obtained by computing the area under the Precision-Recall (PR) curve.

Table 3 presents the results of our proposed work for each part of chicken detection, huddling detection, and inactive chicken detection. Since mAP is an object detector metric, it is only reported for chicken detection. As shown in the table, the chicken detection module accurately detects chickens in the frame with a mAP value of 0.90. For huddling detection, the model can detect 109 out of 123 samples with a precision, recall, and F_1score of 0.93, 0.88, and 0.85, respectively. The false negative samples occur due to occlusion, where the chicken detection module fails to detect occluded chickens, resulting in improper huddling detection. The final part of the model is inactive chicken detection, which is reported in Table 2. The model successfully detects 64 out of 71 inactive chickens, with a precision, recall, and F_1score of 0.92, 0.90, and 0.91, respectively.

Figure 4 showcases the outcomes of chicken detection for four randomly selected samples. Even in challenging environments, the trained model demonstrates its proficiency in accurately detecting chickens within the frame. In samples 1 to 3, featuring images with a multitude of chickens, the model adeptly identifies the majority of them. Sample 4 introduces a new image with distinct lighting conditions. Despite not being trained specifically for this lighting scenario, the model exhibits a high degree of accuracy in detecting the majority of the chickens. Figure 5 presents the outcomes of huddling detection for three randomly chosen samples. The proposed method effectively identifies congregated chickens across various lighting conditions. In Figure 6, the results of inactive chicken detection for two random samples are depicted. The model demonstrates robustness to variations in light conditions and can proficiently identify inactive chickens in diverse environments.

Table 3. Result of the proposed framework

|  | Total | TP | FP | FN | Precision | Recall | F1 score | mAP |
|---|---|---|---|---|---|---|---|---|
| Chicken detection | 2048 | 1145 | 100 | 103 | 0.92 | 0.91 | 0.91 | 0.90 |
| Huddling detection | 123 | 109 | 8 | 22 | 0.93 | 0.88 | 0.85 | - |
| Inactive chicken detection | 71 | 64 | 5 | 7 | 0.92 | 0.90 | 0.91 | - |

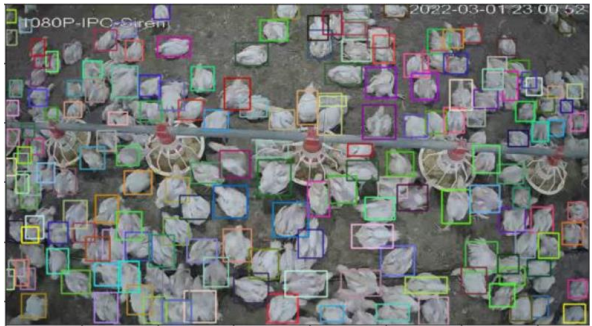
a) sample 1

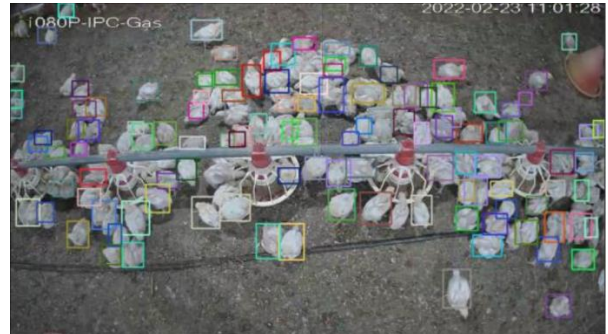
b) sample 2

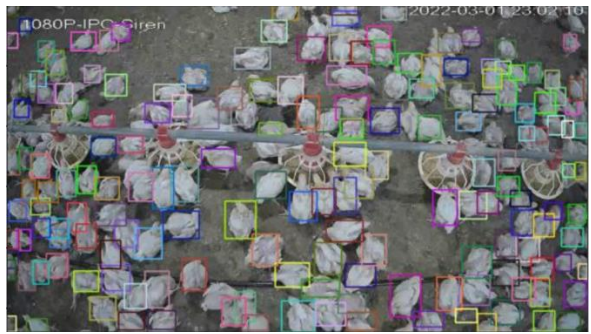
c) sample 3

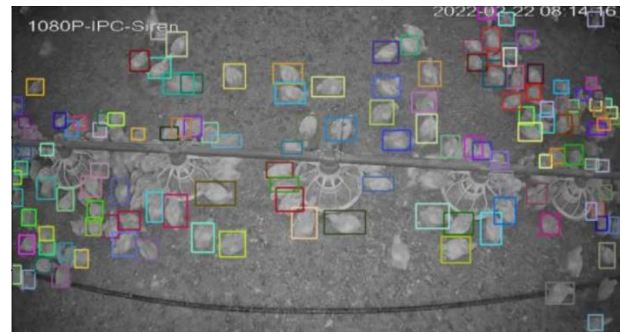
d) sample 4

Figure 5. Chicken detection in four random samples using the trained YOLO model

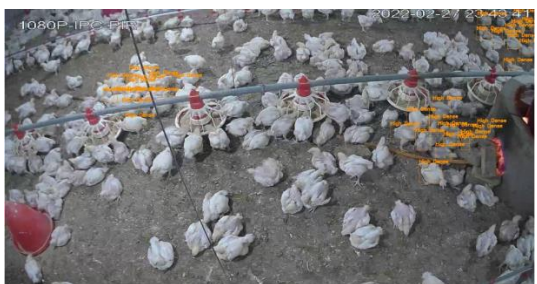
a) sample 1

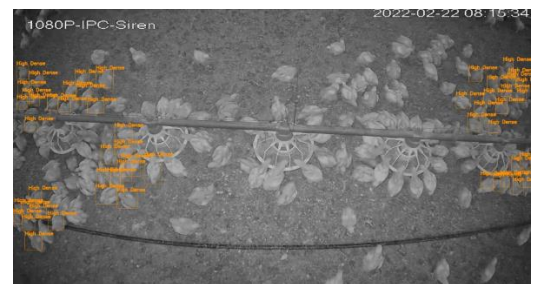
b) sample 2

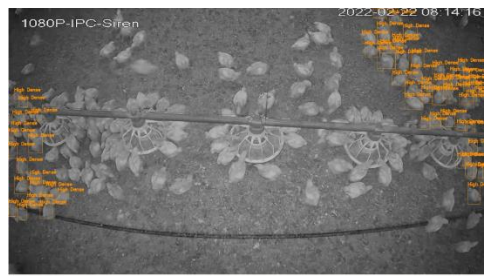
c) sample 3

Figure 6. Huddling detection for three random samples

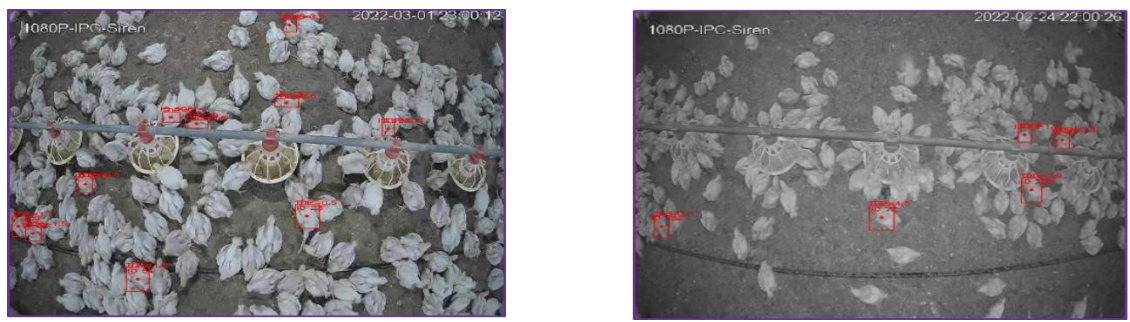

Figure 7. Inactive chicken detection for two random samples

## 4 Conclusion

In conclusion, the integration of AI in poultry farms emerges as a transformative advancement, offering significant benefits. The capability for automated and continuous monitoring of chicken behavior not only facilitates the early identification of potential issues or abnormalities but also contributes significantly to elevated animal welfare and reduced mortality rates on the farm. This, in turn, empowers farmers with the tools for more informed decision-making and the strategic optimization of resources. Our proposed framework for detecting abnormal chicken behavior demonstrates the efficacy of AI applications in this domain. The chicken detection module, achieving an impressive mAP value of 0.90, exhibits exceptional accuracy in identifying chickens within the frame, even in challenging environments. Equally noteworthy, the huddling detection module attains high precision, recall, and F1 score values of 0.93, 0.88, and 0.85, respectively. Similarly, the inactive chicken detection module demonstrates remarkable performance, successfully identifying inactive chickens with precision, recall, and F1 score values of 0.92, 0.90, and 0.91, respectively, regardless of lighting conditions. In totality, our framework consistently delivers reliable performance in the detection and analysis of chicken behavior, highlighting the substantial potential of AI technology in elevating the efficiency and efficacy of poultry farm operations.

## References

[1] Roiter L, Vedenkina I, Eremeeva N. Analysis of the market potential of poultry meat and its forecast. IOP Conference Series: Earth and Environmental Science: IOP Publishing; 2021. p. 022104.
[2] Mohtavipour SM, Shahhoseini HS. A low-cost distributed mapping for large-scale applications of reconfigurable computing systems. 2020 25th International Computer Conference, Computer Society of Iran (CSICC): IEEE; 2020. p. 1-6.
[3] Mohtavipour SM, Shahhoseini HS. GCN-RA: A graph convolutional network-based resource allocator for reconfigurable systems. Journal of Computational Science. 2023;74:102178.
[4] Zhuang X, Bi M, Guo J, Wu S, Zhang T. Development of an early warning algorithm to detect sick broilers. Computers and Electronics in Agriculture. 2018;144:102-13.
[5] Aydin A. Using 3D vision camera system to automatically assess the level of inactivity in broiler chickens. Computers and Electronics in Agriculture. 2017;135:4-10.
[6] Xiao L, Ding K, Gao Y, Rao X. Behavior-induced health condition monitoring of caged chickens using binocular vision. Computers and Electronics in Agriculture. 2019;156:254-62.
[7] Ahmed G, Malick RAS, Akhunzada A, Zahid S, Sagri MR, Gani A. An approach towards IoT-based predictive service for early detection of diseases in poultry chickens. Sustainability. 2021;13:13396.
[8] Mortensen AK, Lisouski P, Ahrendt P. Weight prediction of broiler chickens using 3D computer vision. Computers and Electronics in Agriculture. 2016;123:319-26.
[9] Li G, Zhao Y, Purswell JL, Du Q, Chesser Jr GD, Lowe JW. Analysis of feeding and drinking behaviors of group-reared broilers via image processing. Computers and Electronics in Agriculture. 2020;175:105596.
[10] Ye C-w, Yu Z-w, Kang R, Yousaf K, Qi C, Chen K-j, et al. An experimental study of stunned state detection for broiler chickens using an improved convolution neural network algorithm. Computers and Electronics in

Agriculture. 2020;170:105284.
[11] Fang C, Zhang T, Zheng H, Huang J, Cuan K. Pose estimation and behavior classification of broiler chickens based on deep neural networks. Computers and Electronics in Agriculture. 2021;180:105863.
[12] Mohialdin AM, Elbarrany AM, Atia A. Chicken Behavior Analysis for Surveillance in Poultry Farms. 2023.
[13] Fang C, Huang J, Cuan K, Zhuang X, Zhang T. Comparative study on poultry target tracking algorithms based on a deep regression network. Biosystems Engineering. 2020;190:176-83.
[14] Neethirajan S. ChickTrack–a quantitative tracking tool for measuring chicken activity. Measurement. 2022;191:110819.
[15] Brunet H, Concordet D. Optimal estimation of broiler movement for commercial tracking. Smart Agricultural Technology. 2023;3:100113.
[16] Jaihuni M, Zhao Y, Gan H, Tabler T, Qi H, Prado M. Broiler Mobility Assessment Via a Semi-Supervised Deep Learning Model and Neo-Deep Sort Algorithm. Available at SSRN 4341431.
[17] Li W, Xiao Y, Song X, Lv N, Jiang X, Huang Y, et al. Chicken image segmentation via multi-scale attention-based deep convolutional neural network. IEEE Access. 2021;9:61398-407.
[18] Yang X, Chai L, Bist RB, Subedi S, Wu Z. A deep learning model for detecting cage-free hens on the litter floor. Animals. 2022;12:1983.
[19] Ma X, Lu X, Huang Y, Yang X, Xu Z, Mo G, et al. An Advanced Chicken Face Detection Network Based on GAN and MAE. Animals. 2022;12:3055.
[20] Sozzi M, Pillan G, Ciarelli C, Marinello F, Pirrone F, Bordignon F, et al. Measuring Comfort Behaviours in Laying Hens Using Deep-Learning Tools. Animals. 2022;13:33.
[21] Kodaira V, Siriani ALR, Medeiros HP, De Moura DJ, Pereira DF. Assessment of Preference Behavior of Layer Hens under Different Light Colors and Temperature Environments in Long-Time Footage Using a Computer Vision System. Animals. 2023;13:2426.
[22] Ehsan TZ, Nahvi M, Mohtavipour SM. DABA-net: deep acceleration-based AutoEncoder network for violence detection in surveillance cameras. 2022 International Conference on Machine Vision and Image Processing (MVIP): IEEE; 2022. p. 1-6.
[23] Ehsan TZ, Nahvi M, Mohtavipour SM. An accurate violence detection framework using unsupervised spatial–temporal action translation network. The Visual Computer. 2023:1-21.
[24] Ehsan TZ, Nahvi M. Violence detection in indoor surveillance cameras using motion trajectory and differential histogram of optical flow. 2018 8th International Conference on Computer and Knowledge Engineering (ICCKE): IEEE; 2018. p. 153-8.
[25] Ehsan TZ, Nahvi M, Mohtavipour SM. Learning deep latent space for unsupervised violence detection. Multimedia Tools and Applications. 2023;82:12493-512.
[26] Ehsan TZ, Mohtavipour SM. Vi-Net: a deep violent flow network for violence detection in video sequences. 2020 11th International Conference on Information and Knowledge Technology (IKT): IEEE; 2020. p. 88-92.
[27] Mohtavipour SM, Saeidi M, Arabsorkhi A. A multi-stream CNN for deep violence detection in video sequences using handcrafted features. The Visual Computer. 2022:1-16.
[28] Redmon J, Farhadi A. Yolov3: An incremental improvement. arXiv preprint arXiv:180402767. 2018.
[29] Tan M, Le Q. Efficientnet: Rethinking model scaling for convolutional neural networks. International conference on machine learning: PMLR; 2019. p. 6105-14.
[30] Xie S, Girshick R, Dollár P, Tu Z, He K. Aggregated residual transformations for deep neural networks. Proceedings of the IEEE conference on computer vision and pattern recognition2017. p. 1492-500.
[31] Lin T-Y, Dollár P, Girshick R, He K, Hariharan B, Belongie S. Feature pyramid networks for object detection. Proceedings of the IEEE conference on computer vision and pattern recognition2017. p. 2117-25.
[32] Liu S, Qi L, Qin H, Shi J, Jia J. Path aggregation network for instance segmentation. Proceedings of the IEEE conference on computer vision and pattern recognition2018. p. 8759-68.
[33] Bochkovskiy A, Wang C-Y, Liao H-YM. Yolov4: Optimal speed and accuracy of object detection. arXiv preprint arXiv:200410934. 2020.
[34] Anderson F. Real time, video image centroid tracker. Acquisition, Tracking, and Pointing IV: SPIE; 1990. p. 82.
[35] Ram P, Sinha K. Revisiting kd-tree for nearest neighbor search. Proceedings of the 25th acm sigkdd international conference on knowledge discovery & data mining2019. p. 1378-88.